\documentclass[10pt, a4paper]{article}

\usepackage{lrec2026}
\usepackage{booktabs}
\usepackage{multirow}
\usepackage{graphicx}
\usepackage{amsmath}
\usepackage{url}
\usepackage{microtype}
\usepackage{tcolorbox}
\tcbuselibrary{breakable}
\usepackage{tikz}
\usetikzlibrary{arrows.meta, positioning}

\title{Yale-DM-Lab at ArchEHR-QA 2026: \\
Deterministic Grounding and Multi-Pass Evidence Alignment \\
for EHR Question Answering}

\name{Elyas Irankhah, Samah Fodeh}

\address{Yale University, Department of Emergency Medicine\\
         New Haven, CT, USA \\
         Elyas.irankhah@yale.edu, samah.fodeh@yale.edu}

\abstract{
We describe the Yale-DM-Lab system for the ArchEHR-QA 2026 shared task. The task studies patient-authored questions about hospitalization records and contains four subtasks (ST): clinician-interpreted question reformulation, evidence sentence identification, answer generation, and evidence–answer alignment. ST1 uses a dual-model pipeline with Claude Sonnet 4 and GPT-4o to reformulate patient questions into clinician-interpreted questions. ST2–ST4 rely on Azure-hosted model ensembles (o3, GPT-5.2, GPT-5.1, and DeepSeek-R1) combined with few-shot prompting and voting strategies. Our experiments show three main findings. First, model diversity and ensemble voting consistently improve performance compared to single-model baselines. Second, the full clinician answer paragraph is provided as additional prompt context for evidence alignment. Third, results on the development set show that alignment accuracy is mainly limited by reasoning. The best scores on the development set reach 88.81 micro F1 on ST4, 65.72 macro F1 on ST2, 34.01 on ST3, and 33.05 on ST1.
\\ \newline  }

\begin{document}

\maketitleabstract

%###############################################################################
\section{Introduction}

Electronic health records (EHRs) contain dense clinical narratives. They are difficult for patients to interpret without clinical training. Prior work has studied information extraction from EHR text \cite{liu2013information} and clinical question answering with large language models (LLMs) \cite{singhal2023large, nori2023capabilities}. The ArchEHR-QA shared task \cite{soni-etal-2026-archehr-qa} targets patient-facing access. It asks systems to answer patient-authored questions about a hospitalization episode. The task has four linked subtasks.

A system must reformulate a patient question into a clinician-interpreted question, identify supporting note sentences, generate a grounded answer, and align each answer sentence to its evidence. Evidence grounding is central. Retrieval-augmented generation is a common mechanism for conditioning generation on external text \citep{lewis2020rag}. Citation attribution and faithfulness evaluation provide complementary ways to assess whether model outputs are supported by sources \citep{gao-etal-2023-enabling,honovich-etal-2022-true}.

This paper describes the Yale-DM-Lab submission. Our approach is prompt-based and utilizes in-context learning \citep{brown2020language}. We use self-consistency and voting to improve robustness \citep{wang-etal-2022-self-consistency}. We also evaluate reasoning-style prompting variants when appropriate \citep{wei2022cot}. ST1 uses Claude Sonnet~4 and GPT-4o via their respective direct APIs. ST2--ST4 use Azure-hosted ensembles (o3, GPT-5.2, GPT-5.1, and DeepSeek-R1). We focus on controlled ablations over ensemble size, few-shot count, and vote thresholds.

Our contributions are:
\begin{itemize}
  \item A modular four-subtask pipeline for EHR question answering with a shared Azure infrastructure and consistent few-shot prompting design.
  \item A systematic ablation study evaluating ensemble size, few-shot count, and voting thresholds for evidence identification and alignment.
  \item An analysis of evidence grounding in ST4 showing that providing the full clinician answer paragraph as context improves evidence alignment.
\end{itemize}

%################################################################################
\section{Related Work}

Information extraction from electronic health record (EHR) text has long been studied in clinical natural language processing (NLP)\citet{liu2025ehr_ie}. Early work by \citet{demnerfushman2009cds} developed clinical decision support methods based on structured information extracted from clinical narratives. More recent datasets extend this direction toward question answering. \citet{kweon2024ehrnoteqa} introduced patient-specific QA benchmarks grounded in discharge summaries which enabled evaluation of models on real clinical documentation. Large language models (LLMs) have recently advanced medical question answering. \citet{roberts2020much} demonstrated that large-scale pretrained models acquire substantial factual medical knowledge through language modeling. Building on this progress, \citet{lehman2023clinical} evaluated LLMs in clinical reasoning settings and showed strong performance across medical QA tasks. However, these studies also highlighted limitations when models operate on long clinical documents. EHR-based QA requires reasoning over dense terminology and extended context. Answers must remain grounded in specific source sentences rather than general knowledge. Several studies also address grounding through retrieval mechanisms. \citet{izacard2021leveraging} proposed retrieval-augmented generation (RAG), which conditions generation on retrieved passages and improves factual consistency. In clinical domains, grounding is especially important because hallucinated statements may introduce clinical risk. One study by \citet{ji2023selfreflection} analyzed hallucination behaviors in LLMs and showed that unsupported generations remain a persistent problem. Complementary work by \citet{rashkin2021truthfulqa} examined attribution and truthfulness that emphasized the importance of linking generated claims to supporting evidence. Prompting strategies have been explored to improve reasoning reliability without additional training. \citet{saxena2024evaluating} showed that few-shot prompting with labeled examples improves reasoning consistency even without fine-tuning. \citet{zhou2023least} presented chain-of-thought prompting to encourage intermediate reasoning steps. Extending this idea, \citet{fu2023complexity} proposed self-consistency decoding, which samples multiple reasoning paths and selects the most stable answer. More recently, \citet{kuligin2025prompt} demonstrated that ensembles of diverse models reduce correlated reasoning errors by combining complementary inductive biases. Evaluation of generated clinical text has also evolved. \citet{kryscinski2019neural} showed that lexical overlap metrics alone fail to capture factual correctness in abstractive summarization. \citet{xu-etal-2016-optimizing} introduced SARI to better evaluate paraphrased outputs. Semantic metrics such as BERTScore further improve evaluation of meaning preservation \citet{zhang2019bertscore}. These metrics are particularly relevant for patient-facing clinical answers, which typically paraphrase source text. BLEU therefore provides only limited signal in clinical lay-language generation.

%#########################################################################################

\section{Shared Task and Dataset}

The ArchEHR-QA 2026 dataset contains 167 cases derived from de-identified EHR narratives \cite{soni-demner-fushman-2026-dataset}. Each case includes a patient free-text question, a clinical note segmented into numbered sentences, a clinician-interpreted question, a clinician answer paragraph, and gold evidence alignments. The dataset is split into 20 development cases and 147 test cases. Subtask 1 reformulates the patient question into a clinician-interpreted question of at most 15 words and is evaluated using ROUGE-Lsum, BERTScore, AlignScore, and MEDCON, with the leaderboard score computed as their mean. Subtask 2 identifies the minimal set of supporting sentence IDs and is evaluated using strict micro F1. Subtask 3 generates an answer of at most 75 words grounded in the provided clinical note and is evaluated using BLEU, ROUGE, SARI, BERTScore, AlignScore, and MEDCON. Subtask 4 aligns each clinician answer sentence with supporting note sentences and is evaluated using micro F1. MEDCON is part of the official evaluation for Subtasks~1 and~3 but could not be reproduced locally due to restricted access to the official evaluation environment. Therefore, development experiments reported in this paper use locally reproducible metrics only.
%##################################################################

\section{Methodology}
\label{sec:methodology}

\subsection{Prompting Setup}

We use task-specific prompt templates with fixed instruction blocks. We vary (i) few-shot example count, (ii) model ensemble composition, and (iii) vote threshold and post-processing settings. Few-shot examples are selected in a leave-one-out manner from development cases and are excluded if they match the current case. Unless stated otherwise, prompts enforce strict JSON outputs for structured tasks (ST2, ST4) and task constraints such as word limits and no unsupported facts for generative tasks (ST1, ST3). Across subtasks, the core instruction text remains fixed while experimental variations are applied to configuration parameters. For ST2, we vary few-shot count (3, 10, 19), ensemble members, and merge thresholds. For ST3, we fix the two-stage faithful scaffold and vary few-shot count and model configuration. For ST4, we keep the alignment instruction fixed and vary ensemble and self-consistency settings, vote thresholds, and recall augmentation.

\subsection{Subtask 1 -- Question Reformulation}
\begin{figure}[t]
\centering
\includegraphics[width=\columnwidth]{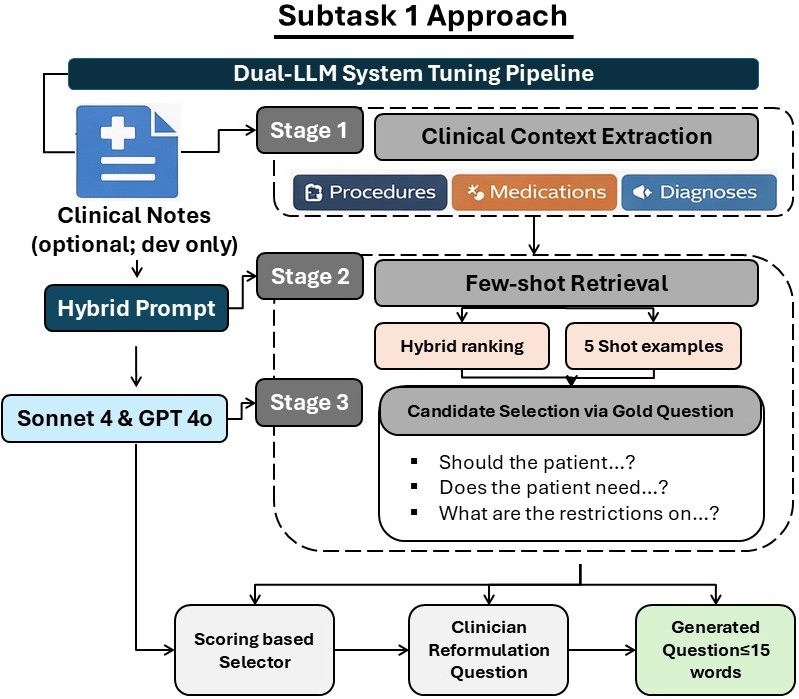}
\caption{Dual-LLM pipeline for Subtask~1 clinician-interpreted question reformulation. The system extracts clinical context, retrieves few-shot examples, generates candidates using parallel LLMs, and selects the final question via scoring-based constraint enforcement. Clinical notes are optional and used only during development; test inference uses patient questions only.}
\label{fig:subtask1_pipeline}
\end{figure}

ST~1 uses Claude Sonnet~4 and GPT-4o via their direct APIs to reformulate the patient question into a $\leq$15-word clinician-interpreted question. We first evaluate a single GPT-4o baseline. We then introduce a dual-model pipeline where Claude Sonnet~4 (Anthropic API) and GPT-4o (OpenAI API) run in parallel. We use:

\begin{itemize}
    \item \textbf{Configurations}: single-model GPT-4o baseline, base dual-model run, and a clinical-reasoning variant with an explicit chain-of-thought step.
    \item \textbf{Dual-model setup}: Claude Sonnet~4 and GPT-4o in parallel; merge by rule selector (pattern match on clinical intent) with fallback to pattern-alignment score.
  \item \textbf{Prompt pipeline (three stages)}: (1)~analyze the clinical note for procedures, medications, findings, and diagnoses to ground the reformulation; (2)~retrieve up to five dev-set examples via hybrid score (question-type + lexical similarity) as few-shot; (3)~score generated candidates with gold question-style patterns and select the highest. Gold questions refer to the ground-truth clinician-interpreted questions from the development set. Candidate questions are scored using pattern alignment heuristics (e.g., question-type matching and lexical similarity to dev gold templates), and the highest-scoring candidate is selected.
  \item \textbf{Hard constraints}: at most 15 words, no first-person pronouns. Figure~\ref{fig:subtask1_pipeline} shows the overall architecture of the Subtask~1 dual-LLM reformulation pipeline.

\end{itemize}

    Clinical note excerpts are used only during development. For the official Subtask~1 test set, note excerpts are not provided. The pipeline therefore operates in a patient-question-only setting, and note-dependent components are not used. All reported test results for Subtask~1 are obtained without clinical notes. Clinical context extraction is a prompt-based preprocessing step that identifies key clinical elements explicitly stated in the inputs. The inputs include the patient question and, during development, an optional note excerpt. Extracted elements include procedures, medications, diagnoses, findings, and temporal or urgency cues. This step is implemented using an LLM instruction prompt, not a separate trained model. The extracted elements guide candidate generation and constrain the reformulation toward the intended clinical information need. Only explicitly stated information is used, and no new clinical facts are introduced.

%############################################################

\subsection{Subtask 2 -- Evidence Identification}

The system prompt instructs the model to return a JSON array of sentence IDs that constitute the minimal evidence set. We explored:

\begin{itemize}
  \item \textbf{Vote threshold}: a sentence is included if it receives $\geq k$ votes across $n$ model-run pairs. Setting $k=1$ (union) maximizes recall; $k = \lceil n/2 \rceil + 1$ (majority) maximizes precision.
  \item \textbf{Few-shot count}: we use 3, 10, or 19 leave-one-out development examples. For each test case, examples are selected from other development cases (excluding the current case) and include the patient question, clinician-interpreted question, and gold evidence sentence IDs. Only cases with non-empty gold evidence are used.
  \item \textbf{Ensemble models}: o3 + GPT-5.2 + GPT-5.1 (baseline trio), with optional addition of DeepSeek-R1.
  \item \textbf{Contrast few-shot}: GOOD examples (correct gold evidence sets) plus BAD examples, e.g.\ one over-inclusive example where the gold set is augmented with a spurious sentence ID so the model sees what to avoid. For instance, a GOOD set may be ["2","5"], while a BAD set may be ["2","5","9"], where sentence "9" is irrelevant. Preference-based methods such as DPO \citep{rafailov2023direct, fodeh2026tabpo} use chosen vs.\ rejected pairs for \emph{training}; we apply the same idea only in-context.

\end{itemize}

The best configuration uses 10 few-shot examples with union merge (min\_votes = 1) and achieves 64.17 strict micro F1 on the development set. Post-processing drops invalid sentence IDs (not present in the case), removes duplicates, and sorts IDs. In the enhanced configuration, we apply additional conservative post-processing (e.g., filtering low-confidence candidates) to improve the score to 65.33.
%#################################################################################

\subsection{Subtask 3 -- Answer Generation}

The prompt gives the patient question, the clinician-interpreted question, the identified evidence sentences (from ST2), and the full note. The model produces a free-text answer, which we hard-truncate to 75 words. We compare:

\begin{itemize}
  \item \textbf{Single model baseline}: o3 only (zero temperature, 5-shot).
  \item \textbf{Ensemble}: o3 + GPT-5.2 + GPT-5.1 with 10 or 15 few-shot dev examples.
    \item \textbf{Faithful variants}: two-stage pipelines that first generate a cited evidence-grounded draft and then rewrite the answer using only the cited sentences. Stage~2 is constrained to avoid unsupported content and not introduce new facts. We evaluate both single-model (GPT-5.2) and ensemble configurations. 
For the ‘Faithful ensemble + BERT selection’ setting, multiple candidate answers are generated from the ensemble, and the final answer is selected by BERTScore-based reranking against the case note text (used as a proxy target).
\item \textbf{Few-shot ablations}: we evaluate 5, 10, and 15 dev examples to study prompt sensitivity.
\end{itemize}

\noindent\textit{In the faithful two-stage setting, Stage~1 produces a draft answer with inline citations (e.g., [12], [15]) that point to the provided evidence sentences. Stage~2 rewrites the answer using only the cited evidence, removes citation markers, and enforces the 75-word limit. This constrains the output to preserve clinical entities and avoids unsupported additions. Example (schematic): Input: question + evidence ([12], [15]). Stage~1: “... [12][15]”. Stage~2: “...”. The final output preserves the same facts and remains supported by the cited evidence.}

%#######################################################################################

\subsection{Subtask 4 -- Evidence Alignment}

\begin{figure}[t]
\centering
\includegraphics[width=\columnwidth]{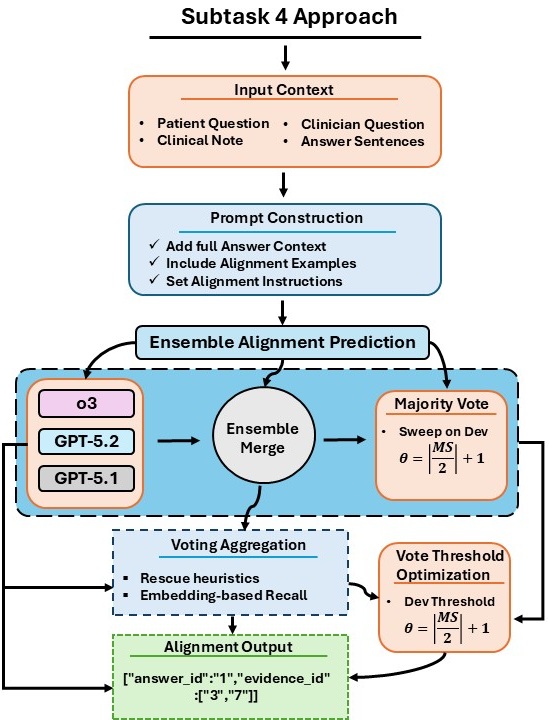}
\caption{Pipeline for Subtask~4 evidence alignment. The system combines ensemble inference, voting aggregation, and embedding-based recall to improve evidence matching.}
\label{fig:subtask4_pipeline}
\end{figure}

ST4 is the most extensively studied subtask as shown in Figure \ref{fig:subtask4_pipeline}. The prompt includes the patient question, the clinician-interpreted question, the full note (numbered sentences), and all answer sentences (numbered). We use:

\begin{itemize}
  \item \textbf{Ensemble and self-consistency}: $M$ deployments, each with $S$ self-consistency samples at $T_s \in \{0.3, 0.4\}$, plus one run at $T=0$. For o3 we use $T=1.0$. A link is kept if its vote count $\geq \theta$, with $\theta = k$ (manual override) or $\theta = \lfloor MS/2 \rfloor + 1$ (majority). Here, $MS$ is the total number of votes ($M$ models $\times$ $S$ self-consistency samples each), we aggregate votes at the $(\text{answer\_id}, \text{evidence\_id})$ link level, keep links supported by more than half of all votes under majority voting, and use $k$ as a manual precision/recall trade-off. 
  \item \textbf{Dev threshold sweep}: On dev we sweep $\theta \in \{1, \ldots, MS\}$, pick the value that maximizes micro F1, write it to \texttt{best\_vote\_threshold.txt}, and reuse it at test time.
  \item \textbf{Full-answer context}: The \texttt{clinician\_answer\_without\_citation} paragraph from the key is added as a ``Full clinician answer (for context)'' block so the model can resolve anaphora and follow the narrative; earlier we used only isolated answer sentences.
    \item \textbf{Embedding-based recall augmentation}: Additional (answer, note) pairs are introduced when their sentence-transformer similarity exceeds a threshold $\tau$ (default 0.68). This mechanism acts as a post-vote recall step to recover missed links after ensemble aggregation. In this work, this step corresponds to the ``rescue heuristics'' reported in the results.
\end{itemize}

%#############################################################
\section{Results}

Table~\ref{tab:all-results} reports performance across all four subtasks. We show development experiments and final test submissions. The experiments evaluate ensemble size, few-shot count, vote thresholds, and grounding strategies.

\paragraph{Subtask 1: Question Reformulation.}
The best development configuration uses note grounding and pattern-based scoring. It achieves a score of 33.05. Dual-model generation improves performance over a single GPT-4o baseline. Chain-of-thought prompting does not provide consistent gains. On the test set, the same configuration achieves 27.09. The decrease reflects the difficulty of producing short clinician-interpreted questions under strict lexical constraints.

\paragraph{Subtask 2: Evidence Identification.}
Evidence identification benefits from ensemble voting. Majority voting increases precision but lowers recall. Union merging increases recall. This improves micro F1. The best development configuration uses 10 few-shot examples with union merging. It achieves 64.17 micro F1. Post-processing increases the score to 65.33. On the test set, the same configuration reaches 61.90 micro F1. These results show that recall-oriented aggregation improves evidence retrieval.

\paragraph{Subtask 3: Answer Generation.}
Answer generation remains difficult. BLEU scores are low. Clinical answers often use varied wording. Ensemble generation improves the overall score compared to single-model generation. The best development score is 34.01. This result uses the o3, GPT-5.2, and GPT-5.1 ensemble. The faithfulness pipeline achieves similar performance. On the test set, the ensemble reaches 30.95.

\paragraph{Subtask 4: Evidence Alignment.}
Evidence alignment produces the strongest results. Ensemble voting improves stability. Embedding-based recall augmentation (rescue heuristics) increases recall. The best development configuration reaches 88.81 micro F1. This setup uses ensemble aggregation with recall-oriented augmentation. Self-consistency sampling provides limited improvement, with performance reaching 83.39 micro F1, which remains below recall-augmented configurations. For example, Table~1 shows $83.39\,\mu$F1 for the self-consistency configuration versus $88.81\,\mu$F1 with embedding-based recall augmentation. On the test set, the ensemble with rescue heuristics achieves 80.41 micro F1.

\paragraph{Observations.}
Several consistent patterns emerge across the experiments. Model diversity provides the most reliable improvement across subtasks. Ensembles outperform single-model configurations and produce more stable predictions. Ensemble disagreement is handled through sentence/link-level vote aggregation. Disagreements occur mainly on borderline evidence, where union-based aggregation preserves recall while majority voting improves precision but may omit relevant sentences. This effect is most visible in ST2 and ST4, where ensemble voting improves robustness and reduces prediction variance. Few-shot prompting also improves performance when examples are well selected. Increasing the number of examples helps until results stabilize. This suggests that example quality is more important than quantity. Aggregation strategy strongly affects evidence retrieval. Recall-oriented approaches perform better than precision-oriented voting. In ST2, union-based merging consistently achieves higher micro F1 than strict majority voting. Missing relevant evidence appears more harmful than including extra candidate sentences, as observed in ST2, where recall-oriented union merging achieves higher micro F1 than precision-oriented majority voting (64.17 vs. 53.40). This effect is not directly measured in ST3 but likely propagates to answer generation through upstream evidence selection. Contextual grounding improves evidence alignment. Providing the full clinician answer paragraph reduces ambiguity and improves matching between answer sentences and supporting note sentences. Evidence alignment remains the most stable task in the pipeline, while answer generation remains the most challenging. Multi-step reasoning continues to introduce errors even for large language models. In ST3, errors arise when combining multiple evidence sentences leads to unsupported or loosely grounded paraphrases. In ST4, errors appear as missing or incorrect alignment links when reasoning fails to consistently map answer statements to all supporting evidence.

\begin{table*}[htbp]
\centering
\small
\setlength{\tabcolsep}{3pt}
\renewcommand{\arraystretch}{1.05}

\begin{tabular}{lcccccc}
\toprule
\multicolumn{7}{c}{\textbf{Development Phase}} \\
\midrule
\multicolumn{1}{c}{\textbf{Method / Config}} & \multicolumn{6}{c}{\textbf{Metrics}} \\
\cmidrule(lr){1-1}\cmidrule(lr){2-7}

\multicolumn{7}{l}{\textit{\textbf{Subtask 1} — Question Reformulation}} \\
& \textbf{Score} & \textbf{R1} & \textbf{R2} & \textbf{RLsum} & \textbf{BERT} & \textbf{Align} \\
\cmidrule(lr){2-7}

Single (GPT-4o) + 5-shot$^{\ddagger}$ 
& 25.49 & 34.67 & 15.28 & 31.76 & 44.71 & -- \\

Dual-model ensemble$^{\ddagger}$ 
& 26.14 & 36.28 & 16.52 & 33.06 & 45.34 & -- \\

Dual + CoT + note grounding$^\dagger$
& 23.98 & 32.33 & 12.48 & 29.19 & 42.74 & -- \\

\textbf{Dual + note grounding + pattern scoring}
& \textbf{33.05} & 46.35 & 29.28 & 44.57 & 54.58 & -- \\

Dual-LLM reasoning 
& 32.14 & 36.28 & 16.52 & 33.06 & 45.34 & 23.85 \\

\midrule

\multicolumn{7}{l}{\textit{\textbf{Subtask 2} — Evidence Identification}} \\

\textbf{} & \textbf{$\mu$P} & \textbf{$\mu$R} & \textbf{$\mu$F1} & \textbf{mP} & \textbf{mR} & \textbf{mF1} \\
\cmidrule(lr){2-7}

10-shot, majority & 72.86 & 42.15 & 53.40 & 55.20 & 41.76 & 45.32 \\
19-shot, majority & 70.83 & 42.15 & 52.85 & 53.32 & 41.66 & 44.53 \\
3-shot, union & 63.55 & 56.20 & 59.65 & 62.40 & 53.84 & 54.07 \\
10-shot, union & 64.71 & 63.64 & 64.17 & 58.85 & 59.55 & 56.77 \\
19-shot, union & 61.54 & 59.50 & 60.50 & 57.12 & 57.33 & 55.33 \\

\textbf{10-shot, union + post-proc}
& 54.75 & 80.99 & 65.33 
& 58.74 & 81.88 & \textbf{65.72} \\

3-shot + contrast few-shot, union & 57.89 & 63.64 & 60.63 & 52.28 & 59.13 & 52.78 \\
4-model (+R1), 10-shot, union & 54.66 & 72.73 & 62.41 & 54.29 & 72.12 & 60.03 \\

\midrule
\multicolumn{7}{l}{\textit{\textbf{Subtask 3} — Answer Generation}} \\

\textbf{} & \textbf{Score} & \textbf{BLEU} & \textbf{RLsum} & \textbf{SARI} & \textbf{BERT} & \textbf{Align} \\
\cmidrule(lr){2-7}

Single (o3), 5-shot & 27.42 & 2.36 & 17.05 & 55.75 & 34.51 & -- \\
Two-stage evidence → answer (GPT-5.2)
& 33.62 & 9.33 & 26.21 & 58.54 & 40.34 & 27.60 \\

Faithful ensemble + BERT selection
& 33.99 & 8.95 & 26.74 & 58.01 & 41.71 & 27.33 \\

\textbf{o3 + GPT-5.2 + GPT-5.1}
& \textbf{34.01} & 9.78 & 26.28 & 58.16 & 39.69 & 28.82 \\
Ensemble + faithfulness, 10-shot
& 33.53 & 9.81 & 25.16 & 59.18 & 39.98 & -- \\
Ensemble, 10-shot & 28.28 & 2.40 & 18.59 & 56.49 & 35.64 & --\\
Ensemble, 15-shot & 27.11 & 2.63 & 17.25 & 53.79 & 34.77 & --\\

\midrule
\multicolumn{7}{l}{\textit{\textbf{Subtask 4} — Evidence Alignment}} \\

\textbf{} & \textbf{$\mu$P} & \textbf{$\mu$R} & \textbf{$\mu$F1} & \textbf{mP} & \textbf{mR} & \textbf{mF1} \\
\cmidrule(lr){2-7}

GPT-5.2 + GPT-5.1, 10-shot & 88.41 & 88.41 & 88.41 & 90.54 & 88.73 & 88.50 \\
\textbf{Embedding recall augmentation}
& 88.49 & 89.13 & \textbf{88.81} & 90.13 & 90.04 & 89.43 \\
Ensemble + Self-consistency (SC) & 84.96 & 81.88 & 83.39 & 87.89 & 84.96 & 85.12 \\
Embedding-only & 65.91 & 42.03 & 51.33 & 74.21 & 44.40 & 52.16 \\

\toprule
\multicolumn{7}{c}{\textbf{Test Phase}} \\
\midrule
\multicolumn{1}{c}{\textbf{Submission}} & \multicolumn{6}{c}{\textbf{Metrics}} \\
\cmidrule(lr){1-1}\cmidrule(lr){2-7}

\multicolumn{7}{l}{\textit{\textbf{Subtask 1} — Question Reformulation}} \\
& \textbf{Score} & \textbf{R1} & \textbf{R2} & \textbf{RLsum} & \textbf{BERT} & \textbf{Align} \\
\cmidrule(lr){2-7}

Dual + note grounding + pattern scoring
& \textbf{27.09} & 30.06 & 10.08 & 28.23 & 40.65 & 19.68 \\
\midrule

\multicolumn{7}{l}{\textit{\textbf{Subtask 2} — Evidence Identification}} \\
\textbf{} & \textbf{$\mu$P} & \textbf{$\mu$R} & \textbf{$\mu$F1} & \textbf{mP} & \textbf{mR} & \textbf{mF1} \\
\cmidrule(lr){2-7}

10-shot, union + post-proc
& 52.34 & 75.71 & 61.90
& 56.68 & 75.52 & \textbf{61.10} \\

\midrule
\multicolumn{7}{l}{\textit{\textbf{Subtask 3} — Answer Generation}} \\
\textbf{} & \textbf{Score} & \textbf{BLEU} & \textbf{RLsum} & \textbf{SARI} & \textbf{BERT} & \textbf{Align} \\
\cmidrule(lr){2-7}

o3 + GPT-5.2 + GPT-5.1
& \textbf{30.95} & 9.11 & 23.13 & 56.54 & 37.24 & 22.07 \\

\midrule
\multicolumn{7}{l}{\textit{\textbf{Subtask 4} — Evidence Alignment}} \\
\textbf{} & \textbf{$\mu$P} & \textbf{$\mu$R} & \textbf{$\mu$F1} & \textbf{mP} & \textbf{mR} & \textbf{mF1} \\
\cmidrule(lr){2-7}

Ensemble + rescue heuristics & 83.30 & 77.70 & \textbf{80.41} & 84.83 & 80.72 & 81.81 \\

\bottomrule
\end{tabular}

\caption{Results across subtasks with Development Phase experiments shown first and Test Phase results summarized at the end. 
$^\dagger$ indicates configurations using explicit reasoning or chain-of-thought(CoT) prompting.
$^{\ddagger}$ indicates baseline configurations without additional grounding or pattern-scoring components.}
\label{tab:all-results}

\end{table*}

%##################################################################################
\section{Conclusion}

We presented the Yale-DM-Lab system for ArchEHR-QA 2026, a four-subtask pipeline for patient-facing EHR question answering. The system combines model-diverse LLM ensembles, self-consistency voting, and leave-one-out few-shot prompting. Our experiments present strong performance across subtasks, with evidence alignment (ST4) reaching 88.81 micro F1 on the development set. The full clinician answer paragraph is used as additional context during evidence alignment. The result emphasizes the critical role of contextual grounding in evidence-based clinical question answering. The implementation of our system, including prompts and inference scripts for all subtasks, is publicly available at \url{https://github.com/Data-Mining-Lab-Yale/ArchEHR2026-CL4Health-LREC}.

\section{Future Work}
Future work will focus on improving data scale, modeling integration, and generation reliability. Larger clinical QA datasets would enable more reliable evaluation and tuning. Joint modeling across subtasks may reduce error propagation. Improving Subtask~3 is a priority, as answer generation remains the weakest component. The task requires producing short, faithful summaries grounded in sparse evidence. We will explore stronger grounding constraints and improved generation strategies. We will further investigate lightweight models that approximate ensemble performance at lower computational cost. This includes fine-tuning on larger EHR corpora and tighter integration between subtasks, such as using ST4 alignments to guide ST3 generation.

%Future work will address these issues. Larger clinical QA datasets would enable more reliable evaluation and tuning. Joint modeling across subtasks may reduce error propagation. We will also focus on improving Subtask~3, which shows the weakest performance among all tasks. Answer generation remains difficult because the model must produce short, faithful summaries grounded in sparse evidence. Future work will explore stronger grounding constraints and improved generation strategies for this step. We also plan to explore lightweight models that can replicate ensemble behavior at lower computational cost. We will also explore fine-tuning on larger EHR corpora and stronger integration between subtasks, such as using ST4 alignments to guide ST3 generation.

\section{Limitations}
This work has several limitations. First, the dataset is small. The development set contains only 20 cases. This limit restricts reliable hyperparameter tuning. It also limits the diversity of few-shot examples. Second, the system relies heavily on prompt engineering. Model behavior depends on prompt wording and example selection. Small prompt changes can alter predictions. This sensitivity reduces reproducibility across deployments. Third, the pipeline treats subtasks mostly independently. Errors can propagate between stages. For example, missing evidence in Subtask~2 may affect answer generation in Subtask~3. The current system does not perform joint reasoning across subtasks. Fourth, the approach depends on large proprietary models. These models require external APIs and significant computational cost. This constraint may limit reproducibility for researchers without access to similar infrastructure.

\section{Ethics Statement}
This work uses de-identified clinical data provided by the shared task. No identifiable patient information is used. The system is intended for research purposes only and not for clinical deployment.
%##################################################################################

\section{Bibliographical References}
\label{sec:reference}

\bibliographystyle{lrec2026-natbib}
\bibliography{languageresource}
\newpage
%############################################################################
\section{Appendix}
\setcounter{figure}{0}
\renewcommand{\thefigure}{A\arabic{figure}}

This section presents the prompt templates used across all four subtasks as (Figure~\ref{fig:prompt_templates}). Each template specifies the instruction structure, role configuration, and output format used during inference. The templates correspond to the prompting strategies described in section~\ref{sec:methodology} and illustrate the constraints applied to each task. Subsections below describe the prompt design for Subtasks~1–4.

\refstepcounter{figure}

\begin{tcolorbox}[
colback=gray!3,
colframe=black,
title= Prompt Templates for Subtasks,
breakable,
label={fig:prompt_templates}
]

\textbf{Subtask 1: Clinician-Interpreted Question}

\begin{tcolorbox}[colback=white,colframe=black]

\textbf{Role:} \texttt{user} only (Claude Sonnet~4 / GPT-4o). Preceded by clinical understanding pre-pass (extracts decisions, conditions, urgency, intent type) and $N$ dev few-shot examples.

\textbf{Core instruction:} Generate 5 clinician-interpreted question candidates. Do NOT add severity/inference words (\textit{low, high, elevated}) unless explicit in patient question. Preserve urgency indicators (\textit{emergency, emergent, salvage}) before the procedure name. Abstract procedural details to clinical intent.

\textbf{Output format:} \texttt{CANDIDATE\_1:}~[q] $\ldots$ \texttt{CANDIDATE\_5:}~[q] \quad ($\leq$15 words, ends \texttt{?}, no first-person. Best selected by pattern-alignment scoring against dev gold templates.

\end{tcolorbox}

\vspace{4pt}

\textbf{Subtask 2: Evidence Selection}

\begin{tcolorbox}[colback=white,colframe=black]

\textbf{Role:} \texttt{user} only (ensemble: o3 + GPT-5.2 + GPT-5.1). 10 dev few-shot examples (leave-one-out).

\textbf{Core instruction:} You are a clinical evidence selector. Given a patient question, a clinician-interpreted question, and numbered note sentences, output the \textit{minimal} set of sentence IDs needed to answer the question.

\textbf{Merge:} union vote (\texttt{min\_votes=1}) across 3 models; invalid IDs dropped post-hoc.

\textbf{Output format:} \texttt{["1","3","7"]} \quad (JSON array of strings; \texttt{[]} if none)

\end{tcolorbox}

\vspace{4pt}

\textbf{Subtask 3: Answer Generation (two-stage faithful)}

\begin{tcolorbox}[colback=white,colframe=black]

\textbf{Role:} \texttt{user} only (cascade: o3$\to$GPT-5.2$\to$GPT-5.1). Using 10 dev few-shot examples.

\textbf{Stage 1} --- cited draft: \textit{``You are a medical expert. Rephrase the relevant note sentences to answer the question. Stay close to the original wording. Cite each used sentence as [2], [7]. 70--75 words. Third person. Do NOT add interpretation beyond the note.''}

\textbf{Stage 2} --- faithfulness rewrite: \textit{``Rewrite the draft using ONLY the cited sentences above. No citation markers in output. Same medical terms and values. 70--75 words.''}

\end{tcolorbox}

\vspace{4pt}

\textbf{Subtask 4: Evidence Alignment}

\begin{tcolorbox}[colback=white,colframe=black]

\textbf{Role:} \texttt{system} + interleaved \texttt{user}/\texttt{assistant} few-shot (up to 20 dev cases, leave-one-out).

\textbf{System:} \textit{``You are a clinical evidence alignment expert. Include a note sentence only when it states the same specific fact, number, or event as the answer sentence (same lab value, dose, procedure, timeline). When in doubt, prefer inclusion (recall matters).''}

\textbf{User turn} adds: 
\texttt{clinician\_answer\_without\_citation} paragraph for full-answer context.

\textbf{Merge:} majority vote $\theta = \lfloor MS/2 \rfloor + 1$; $\theta$ swept on dev gold.

\begin{verbatim}
[{"answer_id":"1",
"evidence_id":["3","7"]},
 {"answer_id":"2",
 "evidence_id":["5"]},
 {"answer_id":"3",
 "evidence_id":[]}]
\end{verbatim}

\end{tcolorbox}

\end{tcolorbox}

\end{document}